\documentclass[runningheads]{llncs}
\usepackage{makeidx}
\usepackage{times}
\usepackage{float}
\usepackage{graphicx}
\usepackage[colorlinks=true, allcolors=blue]{hyperref}
\usepackage{cleveref}
\usepackage{tikz}
\usepackage[bottom]{footmisc}

\usepackage{enumitem}
\usepackage{booktabs}

\usepackage{cite}

\usepackage{listings}
\usepackage{xcolor}
\usepackage{graphicx} 
\lstdefinestyle{promptstyle}{
    backgroundcolor=\color{gray!10}, 
    escapeinside={(*@}{@*)},
    basicstyle=\ttfamily\footnotesize, 
    frame=single,                         
    rulecolor=\color{gray!70},            
    breaklines=true,                      
    postbreak=\mbox{\textcolor{red}{$\hookrightarrow$}\space}, 
    showstringspaces=false,               
    keywordstyle=\color{blue}\bfseries,   
    commentstyle=\color{gray!60}\itshape, 
    stringstyle=\color{teal},             
    numbers=left,                         
    numberstyle=\tiny\color{gray},        
    captionpos=b,   
}

\begin{document}
\mainmatter              

\title{DMN-Guided Prompting: A Framework for Controlling LLM Behavior}
\titlerunning{DMN-Guided Prompting}  
\author{Shaghayegh Abedi\inst{1}  \and Amin Jalali\inst{2}}

\institute{
    Politecnico di Torino, Turin, Italy,\\
    \email{s309894@studenti.polito.it}
    \and
    Stockholm University, Stockholm, Sweden,\\
    \email{aj@dsv.su.se}
}

\maketitle 

\vspace{-0.5\baselineskip}
\begin{abstract}
Large Language Models (LLMs) have shown considerable potential in automating decision logic within knowledge-intensive processes. However, their effectiveness largely depends on the strategy and quality of prompting. Since decision logic is typically embedded in prompts, it becomes challenging for end users to modify or refine it. Decision Model and Notation (DMN) offers a standardized graphical approach for defining decision logic in a structured, user-friendly manner. This paper introduces a DMN-guided prompting framework that breaks down complex decision logic into smaller, manageable components, guiding LLMs through structured decision pathways. We implemented the framework in a graduate-level course where students submitted assignments. The assignments and DMN models representing feedback instructions served as inputs to our framework. The instructor evaluated the generated feedback and labeled it for performance assessment. Our approach demonstrated promising results, outperforming chain-of-thought (CoT) prompting in our case study. Students also responded positively to the generated feedback, reporting high levels of perceived usefulness in a survey based on the Technology Acceptance Model.
\end{abstract}
\vspace{-0.75\baselineskip}	

\vspace{-0.75\baselineskip}
    
\section{Introduction}
\vspace{-0.5\baselineskip}

Large Language Models (LLMs) have shown considerable potential in automating decision logic within knowledge-intensive processes~\cite{khayabashi2025if}. Their capabilities are increasingly leveraged across multiple domains where complex decisions must be made. However, the effectiveness of LLMs critically depends on the quality of their instructions, also known as prompts. Developing such prompts (commonly known as prompt engineering) is often a trial-and-error process, requiring technical expertise and posing challenges related to transparency, reusability, and maintenance~\cite{introduction1,intro2}. As decision logics evolve, their instructional prompts must also be adapted - a task often non-trivial for business users lacking technical proficiency in developing or adjusting prompts.

To address these challenges, we propose a novel framework that integrates Decision Model and Notation (DMN) into prompt template definitions. DMN is a standardized graphical notation for modeling decision logic~\cite{DMN1,DMN2}. By leveraging DMN to modularize and guide prompt structures, our strategy enables LLMs to follow structured decision pathways, breaking down complex logic into interpretable and manageable components. This approach reduces the dependency on ad hoc changes of prompts, making it easier to understand, modify, and maintain - particularly in settings where clarity, consistency, and collaboration are essential.

We implemented our proposed framework as a DMN-guided prompting framework configured in a graduate-level course to automatically support generating feedback for students' assignments while having the instructor in the loop. Students submitted graphical models representing domain-specific decisions. These models were transformed into natural texts, which are given to our framework in addition to instructional DMN rules defined by the course instructor. The framework generated feedback, which was given to students after the instructor labeled them, so students could distinguish between correct and incorrect feedback. These labeled data were used to evaluate the approach, and the results show that DMN-guided prompting outperformed equivalent prompts defined purely using chain-of-thought prompting~\cite{COT}. In addition, students reported high levels of perceived usefulness, as measured through a survey designed based on the Technology Acceptance Model (TAM)~\cite{davis1989perceived}.

The remainder of the paper is organized as follows. Section~\ref{sec:background} provides a brief background on Decision Model and Notation (DMN) and prompt engineering, outlining their relevance for guiding the behavior of Large Language Models. Section~\ref{sec:approach} introduces the proposed DMN-guided prompting framework. Section~\ref{sec:casestudy} presents the case study conducted in a graduate-level course, explaining the implementation process, reporting evaluation results, and discussing the strengths and limitations observed in the study. Finally, Section~\ref{sec:conclusion} concludes the paper and introduces future directions.

\vspace{-0.5\baselineskip}
\section{Background}\label{sec:background}
\vspace{-0.5\baselineskip}

\subsection{Decision Model and Notation}
\vspace{-0.5\baselineskip}

Decision Model and Notation (DMN) is a standard developed by the Object Management Group (OMG) to represent decision logic in a clear, structured, and interpretable format~\cite{DMN2}. It is widely used in business process management to separate decision logic from process flows, enabling organizations to formalize and automate operational decisions transparently. A key advantage of DMN is its accessibility to both business professionals and technical users. It offers graphical and tabular notations that help bridge the gap between informal business rules and executable logic~\cite{dmn3,dmn4}. DMN supports decision modeling using the Decision Requirements Diagram (DRD)~\cite{DMN2}.

\begin{figure}[t!]
    \centering
    \includegraphics[width=1\linewidth]{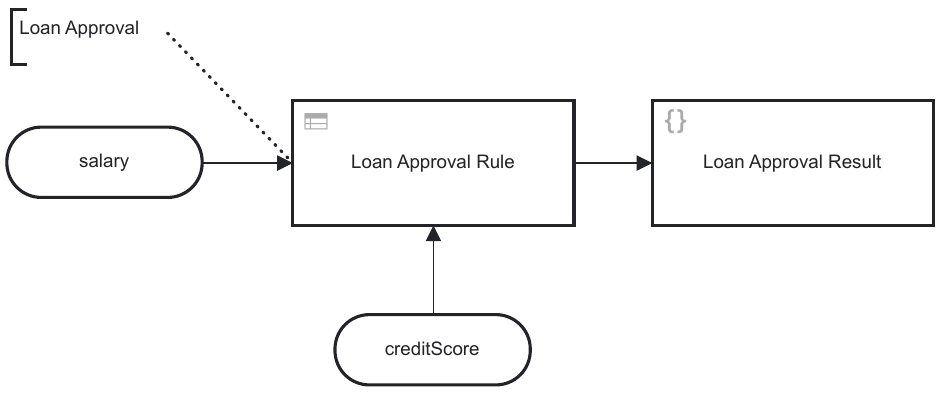}
    \caption{A decision requirements diagram example.}
    \label{fig:dmn_drd}
    \vspace{-1\baselineskip}	
\end{figure}
    
\figurename~\ref{fig:dmn_drd} presents a sample DRD that will serve as a running example throughout this paper to illustrate our approach (for the complete DMN specification, refer to~\cite{DMN2}). This diagram specifies the rules for a fictitious loan eligibility process, where the decision is based on the applicant's \textit{Salary} and \textit{Credit Score}, modeled as \textit{input data} elements. A \textit{decision table} element, named \textit{Loan Approval Rule}, encodes the decision logic. The decision output is then passed to a \textit{literal expression} element, called \textit{Loan Approval Result}. Literal expressions represent decision logic using textual statements - in this case, defining the message to be displayed to the user based on the decision outcome. \textit{Loan Approval} is an annotation that does not have any semantics yet enables providing more information about elements in the DMN model.

\begin{figure}[t!]
    \centering
    \includegraphics[width=0.9\linewidth]{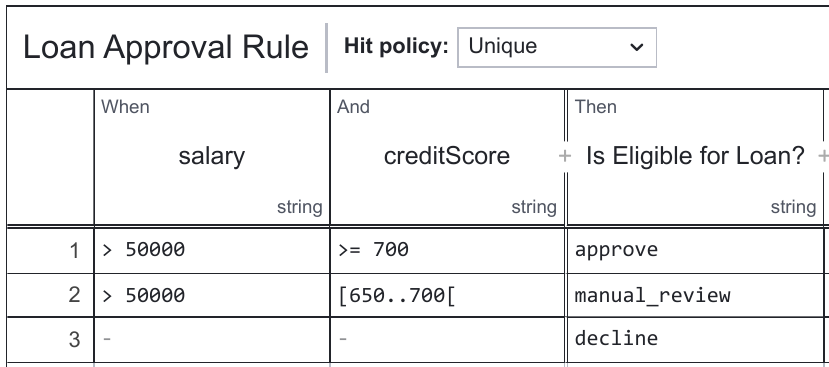}
    \caption{A decision table example.}
    \label{fig:dmn_table}
    \vspace{-1\baselineskip}	
\end{figure}

\figurename~\ref{fig:dmn_table} illustrates how the \textit{Loan Approval Rule} decision table specifies the decision logic. In DMN, a decision table can be used to document the rules governing a decision. Each row in this table represents an individual rule, while columns define the conditions (inputs) and their corresponding outcomes (outputs). This tabular structure is intuitive for many users, resembling familiar tools such as spreadsheets. In our example, the loan will be approved if the applicant’s salary exceeds 50,000 and their credit score is greater than or equal to 700. The rules in this example are evaluated top-down, and the dash symbols in the final row represent a wildcard (``any input value"), which applies when none of the above rules evaluate to true.

\begin{figure}[t!]
    \centering
    \includegraphics[width=1\linewidth]{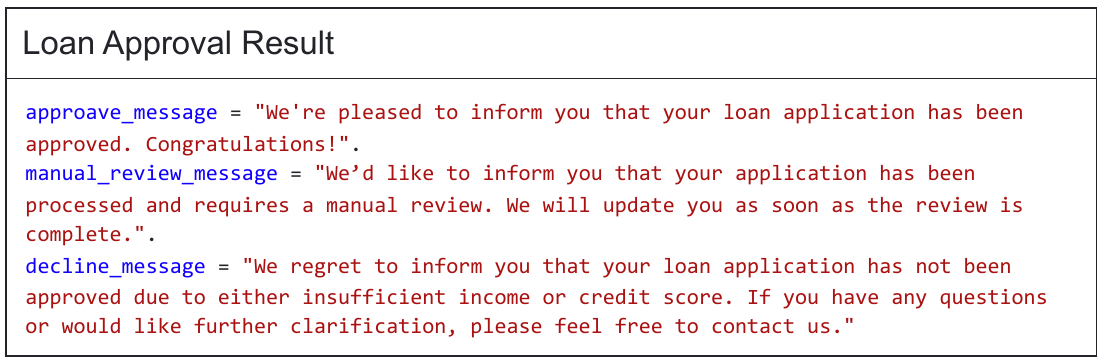}
    \caption{A literal expression example.}
    \label{fig:dmn_literal}
    \vspace{-1\baselineskip}	
\end{figure}

\figurename~\ref{fig:dmn_literal} shows how the \textit{literal expression} is defined using the Friendly Enough Expression Language (FEEL). Based on the decision outcome, this expression determines the message to be shown to the user.

It is important to note that we only present a simplified DMN example here to introduce the essential concepts relevant to this paper. DMN supports a variety of modeling styles, with detailed specifications provided in~\cite{DMN2}. Overall, DMN offers a modular and interpretable framework for defining decision logic. While it has traditionally been applied in operational and compliance domains, its potential for structuring interactions between domain experts and AI-based systems such as Large Language Models (LLMs) remains largely unexplored. This study investigates how DMN can guide LLMs in generating consistent and correct feedback.

\subsection{Prompt Engineering}

Prompt engineering refers to designing input instructions that effectively guide the behavior of Large Language Models. Because LLMs generate output based on probabilistic patterns learned from large datasets, their responses are susceptible to the wording, structure, and context provided in the prompt. As a result, prompt engineering is critical in ensuring the relevance, accuracy, and clarity of the model's output~\cite{prompt_eng1,prompt_eng2}.

A variety of prompting strategies have emerged in recent research. These include zero-shot prompting, where the model is asked to perform a task without examples; few-shot prompting, which provides the model with a few example cases; and Chain-of-Thought (CoT) prompting~\cite{COT}, which encourages the model to generate intermediate reasoning steps before answering. Other techniques, such as retrieval-augmented and role-based prompting, aim to inject external knowledge or frame the model's behavior through personas or task-specific instructions~\cite{prompt_eng3,COT2}.

Among these strategies, Chain-of-thought prompting is reported to be effective for complex tasks as it follows the divide-and-conquer approach, decomposing a task into smaller sub-tasks and helping LLMs to follow a step-by-step approach to solve a problem. For example, instead of prompting the LLM to evaluate an assignment holistically, the task can be divided into sub-prompts that independently assess correctness, clarity, and completeness. These individual judgments are combined to form the final feedback~\cite{dac}. Dividing prompts into individual parts is effective, yet some parts, like business logics, change more frequently than others, requiring the prompt engineer to refine the prompts. A framework enabling the separation of such business rules using DMN can help business people to apply the changes directly, requiring less intervention from a prompt engineer, and enabling organizations to apply changes more easily and quickly.

\vspace{-0.5\baselineskip}
\section{DMN-Guided Prompting}\label{sec:approach}
\vspace{-0.5\baselineskip}

We propose a structured prompting framework in which the behavior of Large Language Models (LLMs) is guided by a formal specification based on the Decision Model and Notation (DMN) standard.

Although DMN supports multiple representations of decision logic, our approach adopts a divide-and-conquer strategy (including chain-of-thought prompting~\cite{COT}) by organizing logic into modular units we refer to as triples. Each triple comprises (1) a set of input data elements, (2) a decision table, and (3) a corresponding literal expression (as illustrated early in the paper in \figurename~\ref{fig:dmn_drd}, including one sample triple). These triples encapsulate logically self-contained decision-making units, each functioning as an independent sub-prompt. This modular structure enables LLMs to process each unit sequentially and coherently.

\begin{figure}[t!]
    \centering
    \includegraphics[width=1\linewidth]{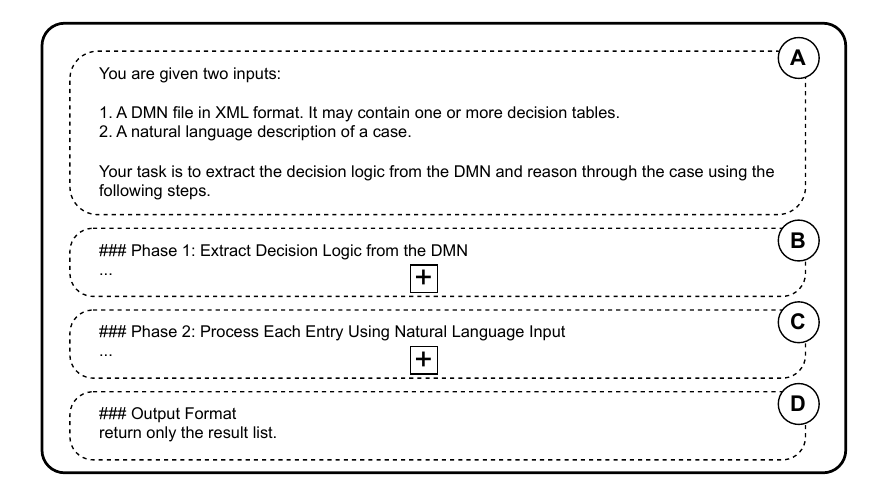}
    \vspace{-0.5\baselineskip}	
    \caption{The overall prompt structure used for DMN-guided evaluation.}
    \vspace{-0.5\baselineskip}	
    \label{fig:prompt}
    \vspace{-1\baselineskip}	
\end{figure}

\figurename~\ref{fig:prompt} outlines the full prompt template used to parse and evaluate the running example DMN file, which can be adopted for different use cases differently. 
Prompt templates refinement is important as it gives organizations freedom in how to design DMN models based on their needs. 
The process is divided into four sequential parts. While the complete contents of Part B and Part C are not shown in the figure, the roles of each part are as follows:
\begin{itemize}[leftmargin=*]
    \item \textit{Part A:} Provides instructional and contextual data, including the DMN file and the input text to be evaluated.
    \item \textit{Part B:} Instructs the LLM to parse the DMN model.
    \item \textit{Part C:} Guides the LLM in evaluating the input text based on the parsed DMN logic.
    \item \textit{Part D:} Directs the LLM to return only the final decision output.
\end{itemize}
In the following sections, we elaborate on \textit{Part B} and \textit{Part C}.

\begin{figure}[t!]
    \centering
    \includegraphics[width=1\linewidth]{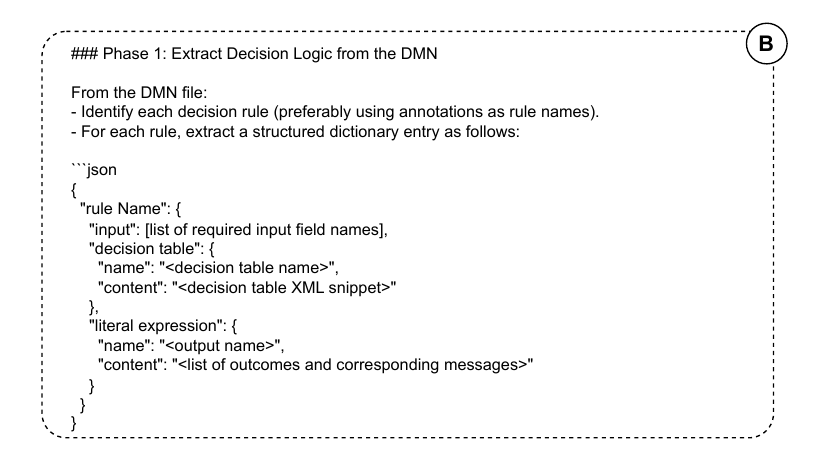}
    \caption{Instructions in Part B: parsing DMN into structured triples.}
    \label{fig:promptB}
    \vspace{-1\baselineskip}	
\end{figure}

\vspace{-0.75\baselineskip}
\subsection{Part B: Parsing the DMN Model}
\vspace{-0.5\baselineskip}
\textit{Part B} instructs the LLM to parse the provided DMN content, as shown in \figurename~\ref{fig:promptB}. The LLM is guided to extract each decision rule using explicit annotations. This approach allows our framework to support multiple decision rules within a single DMN file.
For each rule, the LLM constructs a dictionary containing the following components: rule name, list of input elements, decision table (name and contents), and literal expression (name and contents).
This dictionary forms the knowledge base of triples used to guide subsequent evaluations. Each triple corresponds to a logically encapsulated unit, enabling the LLM to focus on discrete decision points within a larger decision model.

\begin{figure}[t!]
    \centering
    \includegraphics[width=1\linewidth]{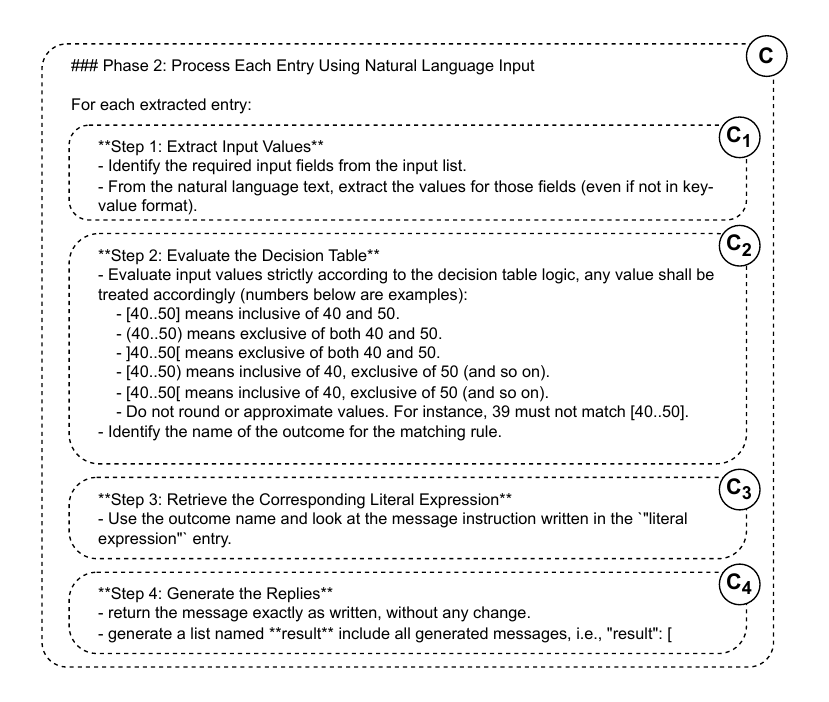}
    \caption{Instructions in Part C: evaluating input text using parsed DMN.}
    \label{fig:promptC}
    \vspace{-1.25\baselineskip}	
\end{figure}

\vspace{-0.75\baselineskip}
\subsection{Part C: Evaluating the Input Text}
\vspace{-0.35\baselineskip}

\figurename~\ref{fig:promptC} illustrates the instructions given in \textit{Part C}, which guide the LLM in evaluating the natural language input using the previously parsed decision logic. This phase comprises four key steps applied to each entry in the extracted dictionary:
\begin{itemize}[leftmargin=*]
    \item \textit{\tikz[baseline=(C.base)]\node[draw,circle,inner sep=1pt](C){\textbf{C1}}; Extract Input Values:} This step instructs the LLM to extract input values from the extracted input list in the extracted dictionary. These input elements guide LLM in extracting their value from the given text (see second instruction line in C1).
    \item \textit{\tikz[baseline=(C.base)]\node[draw,circle,inner sep=1pt](C){\textbf{C2}}; Evaluate Decision Table:} Based on the extracted input values, the LLM evaluates the decision rule using the decision table. Examples are provided to illustrate how to evaluate numeric intervals and enforce exact matching (following a few-shot prompting strategy). These examples were found to be crucial in avoiding hallucinations, particularly in boundary conditions.
    \item \textit{\tikz[baseline=(C.base)]\node[draw,circle,inner sep=1pt](C){\textbf{C3}}; Retrieve Literal Expression:} Once a matching decision rule is identified, the LLM retrieves the corresponding literal expression associated with the outcome.
    \item \textit{\tikz[baseline=(C.base)]\node[draw,circle,inner sep=1pt](C){\textbf{C4}}; Return Final Message:} The final step ensures that only the decision message (from the literal expression) is returned. The prompt instructs the LLM to avoid including intermediate data structures such as the extracted dictionary. Without this constraint, the LLM often included unnecessary reasoning traces, leading to downstream usage confusion.
\end{itemize}

\vspace{-0.5\baselineskip}
\subsection{Result Compilation and Final Output}
After completing all sub-tasks, the LLM assembles the outputs into a single response as described in \textit{Part D}. Each triple is evaluated independently, and the results are synthesized into a coherent final output.
The full implementation, including the Python source code, the running example DMN file, and example input texts, is available on our GitHub repository~\footnote{~\url{https://github.com/Shaghayegh-Abedi/DMN-prompting-project}}. It is important to emphasize that there is no need to modify the prompt if decision rules are changed. Instead, users modify the DMN model, which the LLM interprets to evaluate the logic and generate the results.

Furthermore, if customized message generation is desired, the LLM can post-process the output messages using a secondary prompt. Otherwise, the framework returns only the predefined messages specified in the DMN logic.

This framework offers several key benefits:
\textit{Modularity:} Components can be independently created, reused, or updated, supporting maintainable and scalable prompt design.
\textit{Interpretability:} Each decision rule and its corresponding output are explicit and traceable, enhancing transparency.
\textit{Collaborative design:} Domain experts can contribute directly to changing decision logic and improving quality.
\textit{Transparency:} Most significantly, it enables moving away from ad hoc prompt crafting toward systematic and auditable LLM design for complex, rule-based tasks.

\vspace{-0.5\baselineskip}
\section{Case Study}\label{sec:casestudy}
\vspace{-0.5\baselineskip}

This case study investigates the feasibility and effectiveness of the proposed DMN-guided prompting framework for automated feedback generation in an academic context. Specifically, it examines how decision models can structure the behavior of large language models (LLMs) to evaluate student-created process models and return predefined feedback addressing common known problems in the models. We begin by describing the educational setting, followed by the implementation and evaluation methodology, and conclude with findings concerning system performance and student perceptions. Due to privacy concerns, the student submissions and responses could not be published. 
However, the framework and example materials are provided in our GitHub repository to help readers apply the framework to their own use cases.

\vspace{-0.5\baselineskip}
\subsection{Educational Setting and Study Context}
\vspace{-0.5\baselineskip}
The study was conducted in a graduate-level course titled Business Process Design and Intelligence at the Department of Computer and Systems Sciences (DSV), Stockholm University. A total of 24 student groups participated, each comprising approximately six members. As part of their coursework, students were tasked with designing business process models using Petri nets, implemented via the WoPeD tool~\cite{freytag2005woped} in an agile way~\cite{jalali2018teaching}. The models were expected to reflect best practices in Business Process Redesign (BPR)~\cite{reijers2005best}, such as triage, parallelism, task elimination, automation, and resequencing. Participation in the study was voluntary, and students provided informed consent for using their anonymized submissions in this research.

\vspace{-0.5\baselineskip}
\subsection{Framework Deployment in the Case Study}
\begin{figure}[b!]
    \centering
    \includegraphics[width=0.9\linewidth]{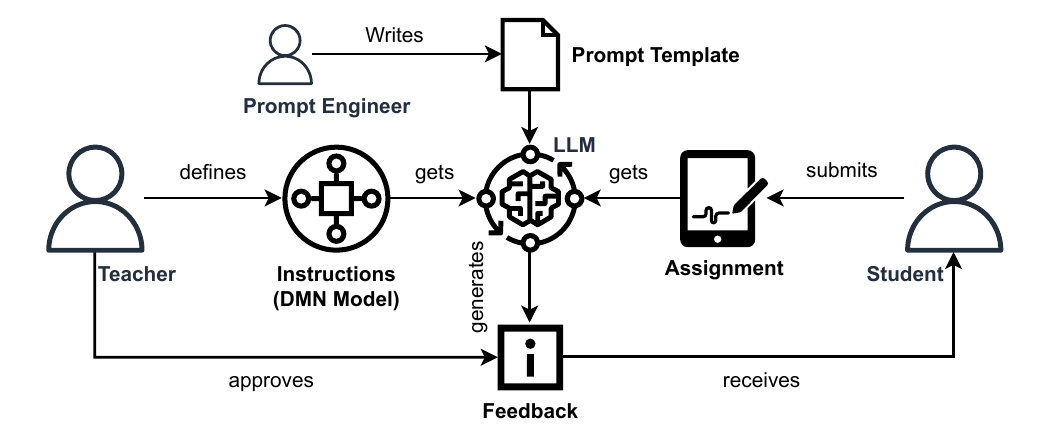}
    \caption{Application of the DMN-Guided Prompting Framework in our case study.}
    \label{fig:method}
    \vspace{-1\baselineskip}	
\end{figure}

\vspace{-0.5\baselineskip}
Figure~\ref{fig:method} depicts deploying the DMN-guided prompting framework within the course. The setup comprises four primary actors: (i) the prompt engineer defining the prompt template, (ii) the instructor, who encodes feedback logic using DMN triples; (iii) the student, who submits process models for evaluation; and (iv) the DMN-guided prompting system, which generates feedback based on the encoded logic for each submission.

To ensure reliability, the generated feedback was not directly shared with students. Instead, a human-in-the-loop mechanism was implemented: instructors manually reviewed each comment and classified it as correct or incorrect. The labeled feedback was subsequently communicated to students, reducing the risk of misleading them into assuming that the received feedback was always correct. This approach safeguarded the instructional quality and enabled an empirical evaluation of the system’s predictive accuracy.
The framework was executed for robustness using two LLMs - GPT-4o (via the OpenAI API) and Gemini 1.5 Pro (via Google’s Gemini API). Additionally, we implemented a Chain-of-Thought (CoT)~\cite{COT} prompting set up as a performance baseline, allowing us to isolate the impact of the DMN structure from general prompt engineering strategies.

\vspace{-0.5\baselineskip}
\subsection{Technical Implementation}
\vspace{-0.5\baselineskip}
We developed a software component to operationalize the framework that translates Petri net models into structured natural language descriptions. These descriptions retain critical control flow and resource semantics, enabling LLMs to interpret process models as standardized textual artifacts. The implementation is publicly available on GitHub for reuse in similar applications.
The core feedback mechanism comprises nine decision rules, each encoded as a table in the Decision Model and Notation (DMN). These rules align with specific BPR principles and provide pedagogically meaningful evaluation criteria for student submissions.

\begin{figure}[b!]
    \centering    \includegraphics[width=1\linewidth]{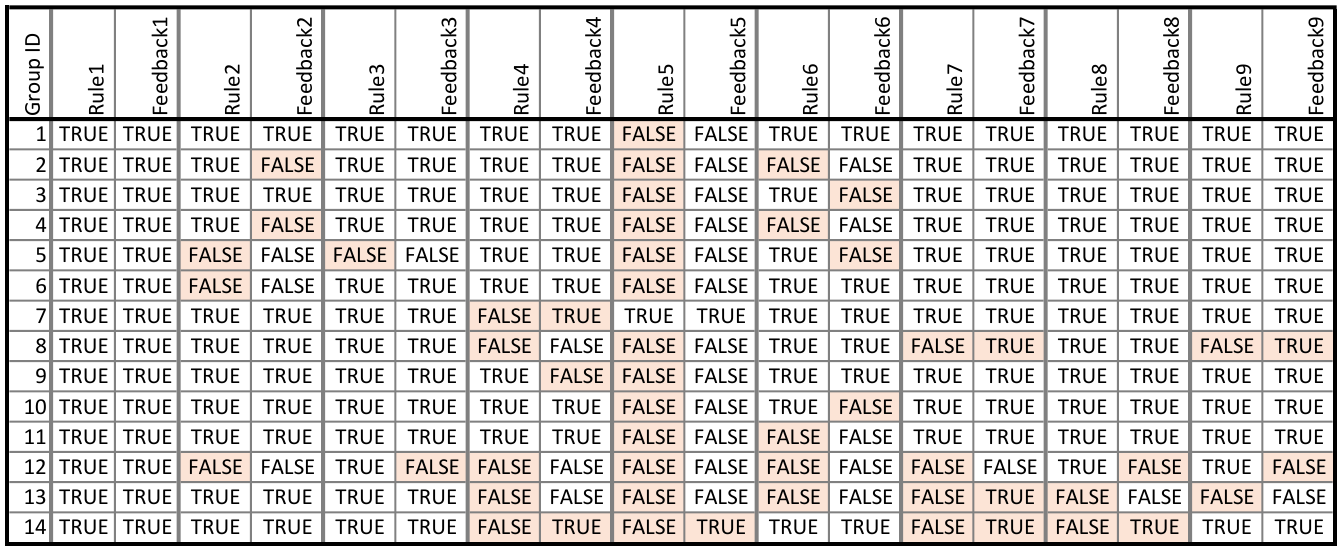}
    \caption{Overview of instructor-labeled evaluations of GPT-4o–generated feedback, presented across student groups and nine BPR rules. Each rule is assessed along two dimensions: whether the student correctly applied the principle (Rule column) and the GPT-4o feedback (Feedback column).}
    \label{fig:labeledfeedback}
    \vspace{-1\baselineskip}
\end{figure}

\vspace{-0.5\baselineskip}
\subsection{Evaluation Results}
\vspace{-0.5\baselineskip}
\figurename~\ref{fig:labeledfeedback} presents an overview of instructor-labeled feedback (for GPT-4o experiment) across student groups and nine BPR rules. Each rule is assessed along two dimensions: the student correctly applied the principle (Rule column), and the LLM-generated feedback (Feedback column). 
\tablename~\ref{tab:comparison_results} provides a comparative analysis of four configurations: DMN-guided and CoT prompting strategies, each paired with GPT-4o and Gemini 1.5 Pro. Standard classification metrics - Precision, Recall, F1-score, and Accuracy - were computed by comparing system outputs with instructor labels.

The specified results in \tablename~\ref{tab:comparison_results} show that the DMN-guided prompting generally outperforms the CoT baseline across most metrics and models.
The GPT-4o + DMN configuration yielded the highest performance: 0.91 precision, 0.90 recall, 0.91 F1-score, and 0.87 accuracy. In contrast, GPT-4o + CoT achieved perfect recall (1.00) but suffered a severe drop in precision (0.36), resulting in a markedly lower F1-score of 0.53 - suggesting a high rate of false positives.

Gemini 1.5 Pro exhibited similar trends. The DMN-guided setup achieved an F1-score of 0.71, outperforming the CoT variant (F1-score: 0.62). These results confirm the value of integrating decision logic into prompt design for enhancing precision, consistency, and pedagogical alignment in automated feedback systems.

\begin{table}[t!]
\centering
\begin{tabular}{|l|c|c|c|c|c|}
\hline
\textbf{Approach} & \textbf{Model} & \textbf{Precision} & \textbf{Recall} & \textbf{F1-score} & \textbf{Accuracy} \\
\hline
DMN-guided & GPT-4o & 0.91 & 0.90 & 0.91 & 0.87 \\
Chain-of-Thought & GPT-4o & 0.36 & 1.00 & 0.53 & 0.54 \\
DMN-guided & Gemini 1.5 Pro & 0.60 & 0.87 & 0.71 & 0.65 \\
Chain-of-Thought & Gemini 1.5 Pro & 0.55 & 0.71 & 0.62 & 0.52\\
\hline
\end{tabular}
\vspace{0.2\baselineskip}	
\caption{Performance comparison of different prompting approaches.}
\label{tab:comparison_results}
\vspace{-1.5\baselineskip}
\end{table}

\vspace{-0.5\baselineskip}
\subsection{Feedback Analysis and Rule-Specific Insights}
\vspace{-0.5\baselineskip}
To complement the quantitative findings, we conducted a detailed qualitative analysis to assess the system’s capability to detect individual BPR principles and reveal underlying student comprehension patterns.
All 14 groups correctly applied Rule 1 (triage), and the DMN-guided framework using GPT-4o flawlessly recognized these instances.

\begin{figure}[b!]
    \centering
    \includegraphics[width=0.7\linewidth]{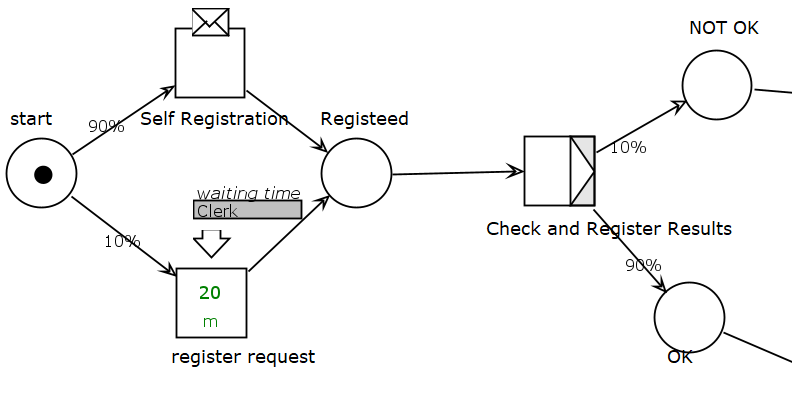}
    \vspace{-1\baselineskip}
    \caption{Example of a merged task with ambiguous labeling, leading to DMN-Guided framework using GPT-4o feedback misclassification for Rule 2 (task composition).}
    \label{fig:rule2}
    \vspace{-1\baselineskip}
\end{figure}

Rule 2 (task composition) was correctly implemented by the majority (79\%) of the groups, and the framework achieved 86\% accuracy in identifying them - closely aligned with the overall accuracy. Notably, in the two misclassified cases, students labeled the merged task using unexpected phrasing (e.g., “check and register result” instead of the expected formulation indicating the task is related to \textit{credit} check), leading to reduced model confidence in mapping the task semantics, as illustrated in Fig.~\ref{fig:rule2}.

For Rule 3 (knock-out), 93\% of the groups modeled the principle correctly, and the framework matched this performance with a 93\% accuracy rate - exceeding its overall average. This rule entails rejecting cases early based on credit score assessments (see the XOR-split after “Check and Register Result” in Fig.~\ref{fig:rule2}).

In contrast, Rule 5 (task resequencing) was the most challenging for students, with only one group (7\%) correctly reordering tasks. Despite this, the framework successfully identified the intended logic in 93\% of the cases. This rule involved moving fraud detection earlier in the process and replacing manual evaluation with a machine learning-based approach. The low implementation rate suggests students struggled with abstract dependency reasoning, signaling a need for more advanced instructional scaffolding.

\begin{figure}[b!]
    \centering
    \includegraphics[width=0.9\linewidth]{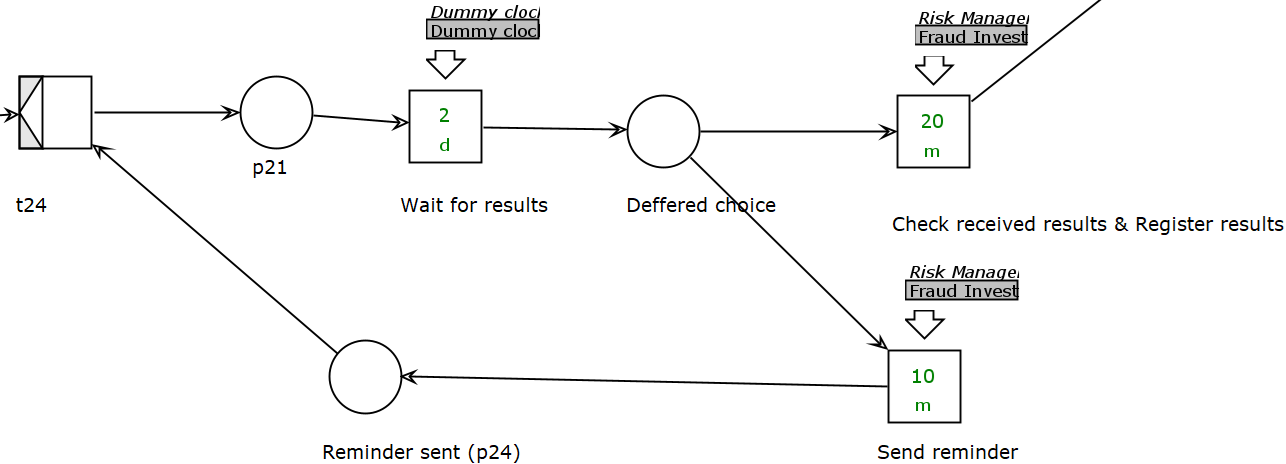}
    \caption{Example of an incorrectly modeled deferred choice, where an unnecessary `Wait for Result' task was added despite the presence of an automated reminder mechanism.}
    \label{fig:rule6}
    \vspace{-1\baselineskip}
\end{figure}

Rule 6 (deferred choice and contact reduction) was correctly applied by 64\% of the groups, with the framework achieving 79\% accuracy. Some groups misused XOR-splits to represent deferred choices. Some groups incorrectly modeled the deferred choice pattern using an XOR-split, which is inappropriate in this context. The task required groups to redesign the process to avoid repeatedly checking the status of an external fraud detection request. According to the business-to-business agreement, the result would be received automatically within two days; otherwise, a reminder should be sent. 

\figurename~\ref{fig:rule6} illustrates an example of a model that does not fully comply with the intended design. The group correctly identified and modeled the deferred choice between sending a reminder and receiving the result (though the automation aspect was not explicitly captured). However, they also added a separate “Wait for Result” task, introducing an unnecessary delay in the process. The DMN-guided framework using GPT-4o successfully identified this issue, even though the merged task (“Receive Result”) was not clearly labeled. This may be due to defining the task to be performed manually by assigning it to the ``Fraud Investigator," which likely helped the LLM interpret the intent from the surrounding context.

Regarding Rule 7 (parallelism between two automated tasks), 71\% of the groups applied it correctly, and the framework reached 79\% accuracy. Some confusion arose when students incorrectly applied task composition - merging two external web services into a single task- violating design control assumptions, as the process designer cannot merge such services.

Finally, for Rules 8 (parallelism of an automated and a manual task) and 9 (sequence within parallelized tasks), 86\% of the groups applied them correctly, and the framework achieved an identical 86\% accuracy - matching its overall performance.

\begin{figure}[b!]
    \centering
    \includegraphics[width=1\linewidth]{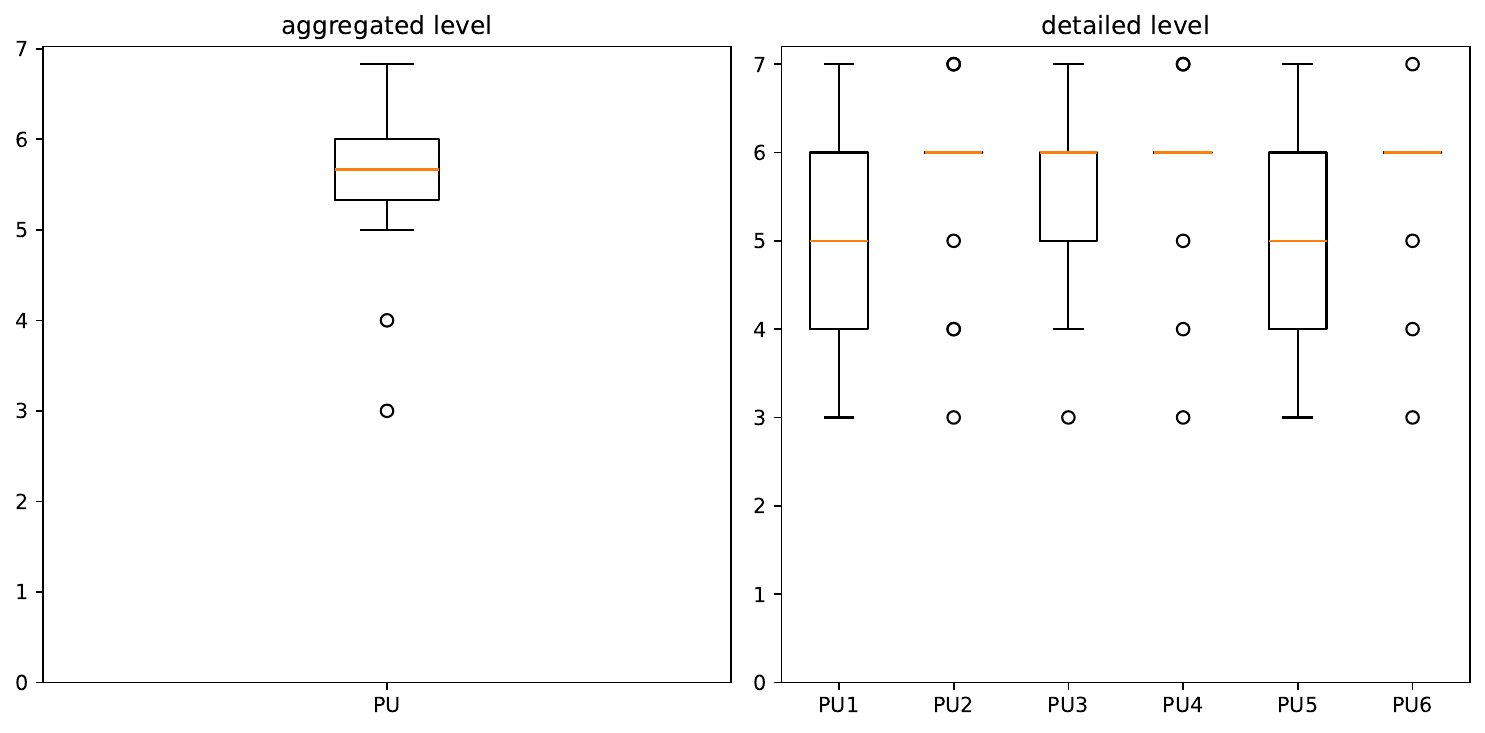}
    \caption{Boxplots showing the distribution of Perceived Usefulness (PU) in aggregated and detailed level.}
    \vspace{-1\baselineskip}	
    \label{fig:tam}
\end{figure}

\vspace{-0.5\baselineskip}
\subsection{Perceived Usefulness}
\vspace{-0.5\baselineskip}
To assess the proposed system's perceived usefulness (PU), we asked students to fill out a survey designed based on the Technology Acceptance Model (TAM)~\cite{davis1989perceived}, a common approach to measure perceived usefulness and ease of use of information systems artifacts~\cite{jalali2021evaluating,jalali2023evaluating,jalali2018hybrid}.  \figurename~\ref{fig:tam} shows the result in aggregated and detailed levels.

At the aggregated level, the overall PU construct, calculated as the average of the six PU items, shows a high central tendency. The median value is above 5.5 on a 7-point Likert scale, indicating strong agreement among participants regarding the system's usefulness. The narrow interquartile range suggests that most responses are concentrated around the upper scale values. While a few outliers exist (e.g., ratings around 3), they do not significantly impact the overall distribution. This confirms that participants, on average, perceived the system to be highly useful.

The detailed boxplots present the distribution of responses for each PU item (PU1 to PU6). All six items reflect consistently high medians, typically at or above 5.5, reinforcing the aggregated findings. Some variation in spread exists across items, particularly for PU5, where responses show a slightly broader distribution. Nonetheless, even in the presence of outliers, the central tendency for all items remains strong, indicating a reliable pattern of positive user perception.

The consistency across items and low variability further support the internal reliability of the PU construct and suggest that the users evaluated the system as useful across different functional dimensions captured by the individual items.

\vspace{-0.5\baselineskip}
\subsubsection{Note on Perceived Ease of Use:}
\vspace{-0.5\baselineskip}
Perceived Ease of Use (PEU) was not assessed, since students interacted with the system only through the university’s Learning Management System (LMS). In this context, PEU would reflect the usability of the LMS rather than our system, making it irrelevant. In contrast, Perceived Usefulness (PU) was evaluated, as students could reflect on the usefulness of the feedback received.

\vspace{-0.5\baselineskip}
\subsubsection{Threats to Validity}
\vspace{-0.5\baselineskip}
Our study has several limitations. \textit{Internal validity} may be affected by the manual labeling of feedback, which introduces subjectivity despite the human-in-the-loop review. \textit{Construct validity} is constrained by the chosen metrics (precision, recall, F1, accuracy) and the survey-based perceived usefulness, which may not fully capture long-term learning outcomes. \textit{External validity} is limited since the case study was conducted in a single graduate-level course with 24 groups and nine decision rules, reducing generalizability to other domains and contexts. Finally, \textit{conclusion validity} may be influenced by the relatively small dataset and ceiling effects for some rules. Further studies in diverse educational and professional settings are needed to strengthen the generalizability of our findings.

\vspace{-0.5\baselineskip}
\section{Conclusion}\label{sec:conclusion}
    \vspace{-0.7\baselineskip}

This paper presented a novel prompting framework that integrates Decision Model and Notation (DMN) with Large Language Models (LLMs) to guide decision logic in a modular, interpretable, and user-friendly manner. By structuring prompts as DMN triples (each consisting of input data, decision tables, and literal expressions), our approach operationalizes a divide-and-conquer strategy for LLM behavior control. The case study conducted in a graduate-level course demonstrated that DMN-guided prompting generally outperformed chain-of-thought (CoT) methods in predictive accuracy. Regarding user-perceived usefulness, students reported high perceived value from the feedback generated by our DMN-based approach.

The results confirm the viability of DMN for LLM control in educational settings, particularly for structured feedback generation. More broadly, this work illustrates the untapped potential of DMN to encode domain-specific reasoning in a format that LLMs can interpret and execute reliably, offering a promising direction for rule-based, transparent AI systems.

Future research can expand in several directions. First, exploring how DMN-guided prompting can be adapted for other application domains, such as healthcare or compliance auditing, will help assess its generalizability. Second, combining DMN with retrieval-augmented generation (RAG) may further enhance reasoning grounded in external knowledge. Third, if customized message generation is desired, the LLM can post-process the output messages using a secondary prompt; otherwise, the framework returns only the predefined messages specified in the DMN logic. Fourth, this approach can be complemented by incorporating few-shot learning, where example inputs are used to improve LLM accuracy when interpreting or applying DMN rules. Fifth, other ways of defining DMN rules and alternative strategies for prompt construction can be explored to increase flexibility and expressiveness. Finally, investigating multi-turn interactions and conversational agents powered by DMN-guided LLMs could open new opportunities for explainable and controllable AI interfaces in decision support systems.


\subsubsection{Acknowledgment:}
The computations for testing and developing our approach were enabled by resources provided by the National Academic Infrastructure for Supercomputing in Sweden (NAISS), partially funded by the Swedish Research Council through grant agreement no. 2022-06725.
\vspace{-0.7\baselineskip}

\bibliographystyle{abbrv} 
\bibliography{main} 

\appendix
\section{Appendix}

This section provides a list of questions asked in the survey to measure perceived usefulness, following the Technology Acceptance Model (TAM). 
Participants rated their agreement with the following statements on a 7-point Likert scale (1 = Strongly Disagree, 7 = Strongly Agree):
\begin{enumerate}
	\item Using the auto-feedback system enabled me to identify relevant BPR patterns more quickly.
	\item Using the auto-feedback system improved my performance in applying BPR patterns to business process redesign.
	\item Using the auto-feedback system increased my productivity when analyzing and revising business process solutions.
	\item Using the auto-feedback system enhanced my effectiveness in recognizing missing BPR patterns.
	\item Using the auto-feedback system made it easier for me to understand which BPR patterns were applicable.
	\item  I found the auto-feedback system useful for learning and applying BPR patterns in business process redesign tasks.	
\end{enumerate}

\end{document}